\pdfoutput=1

\documentclass[11pt]{article}
\usepackage{booktabs}
\usepackage{graphicx}
\usepackage{algpseudocode}
\usepackage[preprint]{acl}
\usepackage{multirow}
\usepackage{enumerate}
\usepackage{amsfonts}

\usepackage{latexsym}
\usepackage{xspace}
\usepackage{amsmath}

\usepackage[T1]{fontenc}
\usepackage{pifont}
\usepackage{amssymb}
\usepackage{newfloat}
\usepackage{listings}
\usepackage{multirow}
\usepackage{booktabs}
\usepackage{times}
\usepackage{latexsym}
\usepackage[T1]{fontenc}    
\usepackage[utf8]{inputenc}
\usepackage[inline]{enumitem}
\usepackage[dvipsnames]{colortbl}
\usepackage[utf8]{inputenc}
\usepackage[T1]{fontenc}
\usepackage{babel} 
\usepackage[toc,page]{appendix}
\usepackage{makecell}
\usepackage{soul}
\usepackage{verbatim}
\usepackage{arydshln}
\usepackage{amssymb}
\usepackage{graphicx}

\usepackage{hyperref}

\usepackage{times}
\usepackage{latexsym}

\usepackage[T1]{fontenc}

\usepackage[linesnumbered,ruled,vlined]{algorithm2e}
\usepackage{adjustbox}
\usepackage{footmisc}
\usepackage{microtype}
\usepackage{bbm}
\usepackage{wasysym}
\usepackage{tikz}
\usepackage{tcolorbox}
\usepackage{siunitx}
\usepackage{pifont}
\usepackage{longtable}
\usepackage{supertabular}
\usepackage[fixed]{fontawesome5}

\usepackage[utf8]{inputenc}

\usepackage{microtype}
\usepackage[nameinlink]{cleveref}

\usepackage{inconsolata}

\usepackage{graphicx}

\newcommand{\ie}{\emph{i.e.,}\xspace}
\newcommand{\aka}{\emph{a.k.a.,}\xspace}
\newcommand{\eg}{\emph{e.g.,}\xspace}

\newcommand{\ignore}[1]{}

\definecolor{backred}{RGB}{255, 190, 190}
\definecolor{backblue}{RGB}{220, 230, 250}

\title{What Affects the Stability of Tool Learning? An Empirical Study on the Robustness of Tool Learning Frameworks}

\author{
Chengrui Huang$^{1*}$, Zhengliang Shi$^{2}$\thanks{The first two authors contributed equally and the order is alphabetical}, Yuntao Wen$^{1}$,\\
{\bf Xiuying Chen$^{3}$, Peng Han$^{1}$, Shen Gao$^{1}$\footnotemark[2], Shuo Shang$^{1}$} \\
$^{1}$ University of Electronic Science and Technology of China,\\
$^{2}$ Shandong University,
$^{3}$ Mohamed bin Zayed University of Artificial Intelligence\\
\texttt{yunrongyuxi@gmail.com},
\texttt{shizhl@mail.sdu.edu.cn},\\
\texttt{yuntaowenx@gmail.com},
\texttt{xiuying.chen@mbzuai.ac.ae},\\
\texttt{penghan@uestc.edu.cn},
\texttt{shengao@pku.edu.cn},
\texttt{jedi.shang@gmail.com}
}

\newtcbox{\hlprimarytab}{on line, rounded corners, box align=base, colback=c3!10,colframe=white,size=fbox,arc=3pt, before upper=\strut, top=-2pt, bottom=-4pt, left=-2pt, right=-2pt, boxrule=0pt}
\newtcbox{\hlsecondarytab}{on line, box align=base, colback=red!10,colframe=white,size=fbox,arc=3pt, before upper=\strut, top=-2pt, bottom=-4pt, left=-2pt, right=-2pt, boxrule=0pt}

\definecolor{Gainsboro}{rgb}{0.86, 0.86, 0.86}

\definecolor{Gray}{gray}{0.95}
\definecolor{LightCyan}{rgb}{0.88,1,1}

\begin{document}
\maketitle
\begin{abstract}
Tool learning methods have enhanced the ability of large language models (LLMs) to interact with real-world applications. 
Many existing works fine-tune LLMs or design prompts to enable LLMs to select appropriate tools and correctly invoke them to meet user requirements.
However, it is observed in previous works that the performance of tool learning varies from tasks, datasets, training settings, and algorithms.
Without understanding the impact of these factors, it can lead to inconsistent results, inefficient model deployment, and suboptimal tool utilization, ultimately hindering the practical integration and scalability of LLMs in real-world scenarios.
Therefore, in this paper, we explore the impact of both internal and external factors on the performance of tool learning frameworks. 
Through extensive experiments on two benchmark datasets, we find several insightful conclusions for future work, including the observation that LLMs can benefit significantly from increased trial and exploration. 
We believe our empirical study provides a new perspective for future tool learning research.
\end{abstract}

\section{Introduction}\label{sec:intro}

Tool learning aims to augment LLMs with external tools, teaching them how to select appropriate tools, generate correct parameters and ultimately parse execution results to produce correct responses~\citep{toolw,api-bank, toolformer}.
By learning to use various tools, LLMs can better assist users in completing practical tasks, such as planning itineraries~\cite{xie2024travelplanner}, controlling physical robots~\cite{wang2023voyager} and accessing the Web~\cite{webcpm}. 
This capability is crucial for enhancing the interaction between LLMs and real-world applications, empowering them as agents to provide more comprehensive and useful assistance~\cite{chameleon,liu2023agentbench,tian2024opportunities}.

\begin{figure}[!t]
\centering
\includegraphics[width=1.0\linewidth]{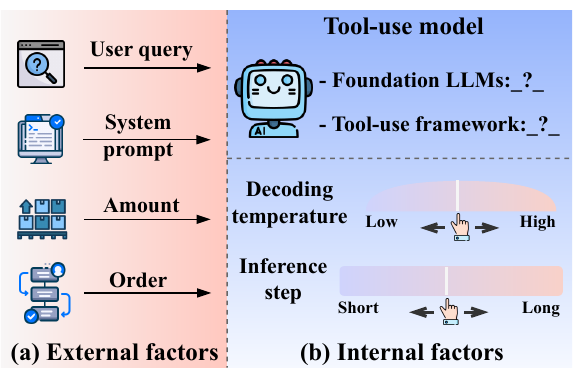}
\caption{Illustration of various factors that may affect the robustness of tool learning methods.}
\label{fig:intro} 
\end{figure}

In the tool learning tasks, most of previous work focuses on improving the performance of LLMs in successfully solving complex tasks, including designing chain-of-thought framework~\cite{chameleon}, employing multi-agent algorithms~\cite{shi2024learning,qiao2024autoact} or tuning  models on specific tool-use datasets~\cite{toolalpaca},
Numerous empirical studies are also conducted to evaluate the tool-use capability of LLMs, such as when to use, how to use, and which tool to use~\cite{xu2023tool,huang2023metatool}.
Despite their progress, we find that \textit{stability}, a crucial dimension to reflect the performance variation of LLMs under volatile scenarios~\cite{li2023robust,gu2022robustness}, is less investigated.
In real-world applications, various factors can affect the performance of tool learning models, and sometimes even produce different responses to identical user queries, \aka instability.
For example, ~\citet{ye2024rotbench} show that even simple perturbations can cause models to select entirely incorrect tools or generate incorrect tool-calling parameters. 
These seemingly unrelated perturbations can lead to the failure of the task.
Therefore, comprehensively exploring the factors related to the stability issue and quantitatively analyzing their impact becomes necessary for practical scenarios.

In this work, we provide the first empirical study on systematically analyzing the stability of tool-use models. 
To achieve this, we first categorize the diverse factors into two categories: \textbf{internal} and \textbf{external} factors.

The \textbf{internal} factors indicate uncertainties during the development of tool-use models from the developers' perspective. 
As shown in Figure~\ref{fig:intro}, we consider the decoding temperature, the maximum inference steps, and the selection of different foundation LLMs. 
Given the numerous works that guide LLMs to automatically use external tools~\cite{react, yang2023mm, toolllm}, we also analyze the impact of different tool-use frameworks on the model's performance.
Exploring these internal factors will help enhance the performance of the framework during development. 
Different from the internal factors, \textbf{external} factors primarily involve diverse prompt engineering when interacting with established tool-use models, which are beyond the control of developers once the models are deployed. 
Specifically, these factors includes different styles of user queries, customized system prompts for tool-use models, and the candidate toolset used to solve a query. 
For a holistic investigation, we change the candidate toolset by reordering it or expanding its scale, respectively.
Investigating these external factors will help developers understand the stability in user-facing scenarios, thereby improving the overall user experience.

To quantitatively validate the impact of the aforementioned internal and external factors on the tool learning process, we conduct extensive experiments on the most commonly used ToolBench~\cite{toolllm} dataset.
We employ several commonly used metrics to measure the performance from multiple perspectives and derive a series of interesting findings.
We highlight the following:
\begin{itemize}
    \item Existing tool-use workflow exhibits obvious instability towards various internal and external factors. Even the state-of-the-art methods still exhibit instability with inessential perturbations.
    \item Among the internal factors, the proper hyper-parameter settings may boost the LLMs to generate diverse solutions. However, it also leads to instability. 
    \item Among the external factors, the LLMs are sensitive to the change of candidate toolset (\ie order or scale) and the system prompts.
    \item The advanced tool selection algorithms (\ie tree-based search) can improve the accuracy, but they may suffer from accumulated hallucination with less stability, as well as substantial inference costs.
\end{itemize}

\section{Related work}\label{sec:related}

\paragraph{Tool learning with LLMs.}
Tool learning aims to augment LLMs with real-world tools,  extending their utility and empowering them as agents to automatically solve practical tasks~\cite{toolw,toolalpaca,shi2023towards,gao2024confucius}. 
Pioneering work like Toolformer~\citep{toolformer} and ToolkenGPT~\cite{toolkengpt} teaches LLMs to utilize tools by training on specific tool-use datasets~\cite{patil2023gorilla,wang2024executable}.
Recent work leverages the inherent in-context learning capability of LLMs to master various tools, where the demonstration and usage are taken as the prompt \citep{yang2023mm,shi2024chain,guo2024large}.
Despite the progress of recent tool-use models in successfully solving complex tasks, their stability is less investigated.
In this work, we provide a comprehensive empirical study on the stability of them across diverse scenarios.

\paragraph{Evaluation of tool-use LLMs.}

In tool learning tasks, previous work primarily evaluates the success rate of LLMs in completing tasks, such as Success Rate~\cite{gpt4tools, restgpt} and Win Rate~\cite{toolllm}.
Recently, the ToolSword~\cite{ye2024toolsword} has also proposed to unveil safely-related issues of LLMs during the tool learning process.
However, stability, a crucial dimension related to practical applications~\cite{wang2023robustness}, has been less investigated. 
Although some work, like RotBench~\cite{ye2024rotbench}, proposes evaluating the robustness of tool-use LLMs, they only consider the different types of noise injected into original candidate toolsets. 
To the best of our knowledge, a thorough stability evaluation of tool-use LLMs remains under-explored.
In our work, we fill this gap by providing a systematic evaluation of the stability of tool-use LLMs, and quantitatively analyzing their drawbacks under different settings.

\section{Experimental Settings}\label{sec:dataset-evaluation}

\subsection{Dataset}\label{sec:dataset}

We conduct experiments on the subset of widely-used ToolBench~\cite{toolllm} benchmark, including \textit{I1-instruction} and \textit{I1-tools}.
Each dataset contains 200 tasks involving various real-world applications, which evaluates tool-use models under practical scenarios.
The detailed statistics can be found in Table~\ref{tab:dataset}.

The original ToolBench only provides a task-solving trajectory of GPT-3.5 as an evaluation reference, which includes both ground truth and irrelevant tools. 
However, commonly used evaluation metrics (\S~\ref{sec:evaluation}) require computing the overlap between model-selected tools and the ground truth. 
Therefore, we repurpose ToolBench to support our evaluation. 
For each task, we extract the tools used in the original solution. 
Next, we invite three well-educated experts with relevant research backgrounds to manually select the correct tools for solving the task. 
Several strategies are employed to ensure the quality of this process, which can be found in Appendix~\ref{sec:app:repurpose}.

\begin{table}[!t]
    \centering
    \resizebox{\linewidth}{!}
    {\Large
    \begin{tabular}{lcccc}
    \toprule
         \textbf{\# Dataset}&  \textbf{\# Amount} &\textbf{\# Category} &\textbf{\# APIs} &\textbf{\# Avg. APIs} \\ \midrule
         I1-inst.& 200& 36& 995& 5.34 \\ 
         I1-tool& 200 & 33 & 548 & 4.79 \\ \bottomrule
     \end{tabular}
     }
    \caption{Statistics of the experimental datasets such as the count of task and tool category. The \textit{Avg. APIs} indicates the average of the candidate toolset per task.}
    \label{tab:dataset}
\end{table}

\begin{figure}[!t]
        \centering
	\includegraphics[width=1
 \linewidth]{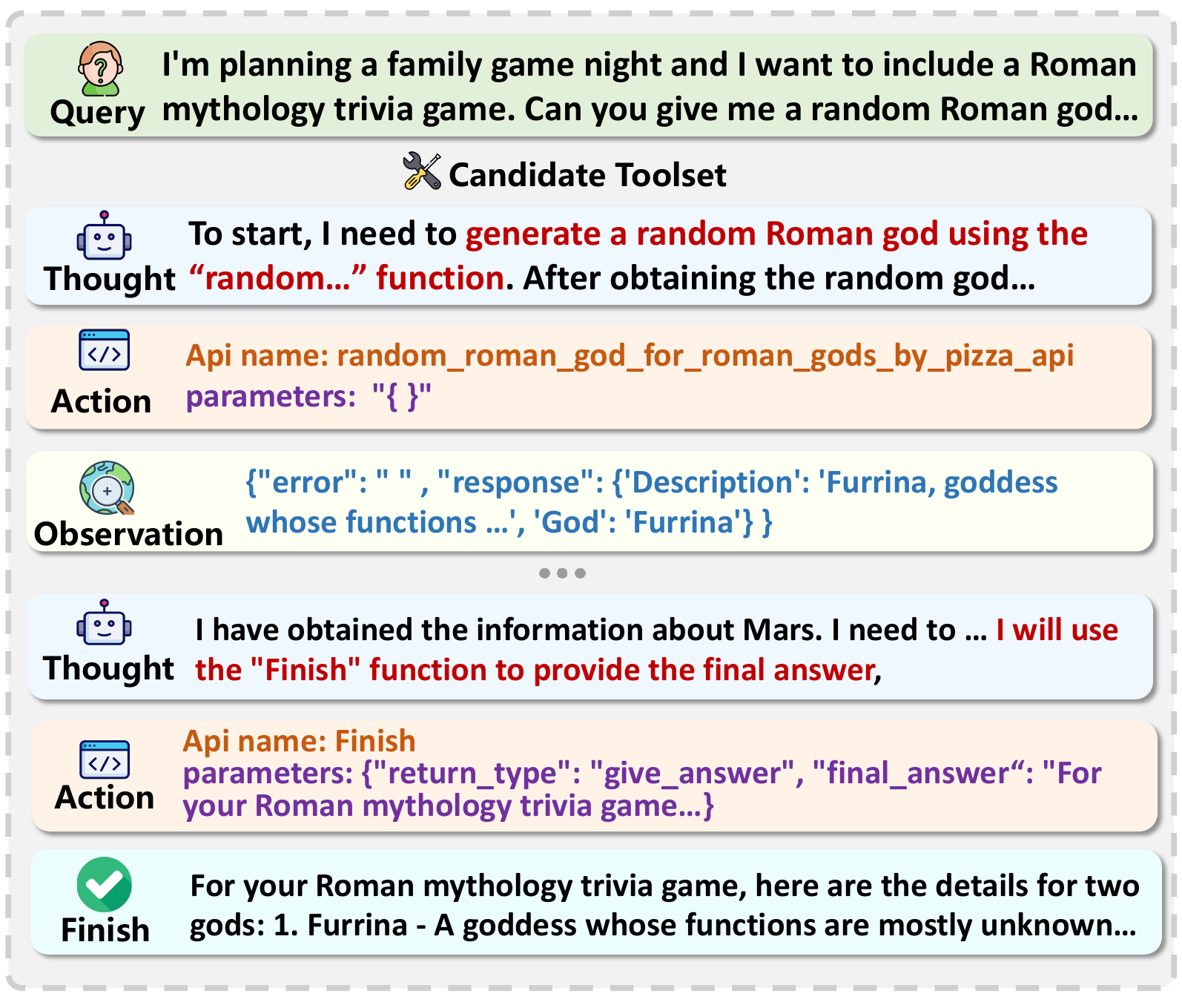}
        \caption{ The default tool-use framework in our work. The LLM is guided to iteratively decide which tool to use (\textit{Thought}), execute the selected tool (\textit{Action}), and incorporate the execution results into context (\textit{Observation}) for the next iteration prediction. }
 \label{fig:react}
\end{figure}

\subsection{Evaluation Metrics}\label{sec:evaluation}

Following previous work~\cite{ye2024rotbench, restgpt}, we use the \textit{Success Rate} and \textit{T-test} as evaluation metrics.
We also consider the \textit{Give Up Rate}, \textit{Invalid Selection Rate} as metrics for a comprehensive evaluation.

\paragraph{Success Rate (Success\%).}
This metric intuitively evaluates the capability of tool-use LLMs in correctly selecting tools and generating corresponding arguments for execution.
It calculates the proportion of tasks that the model can complete successfully within limited inference steps.
The success rate is 1 if and only if all the required tools are used to solve a task.

\paragraph{T-test.}
To analyze the stability of tool-use LLM towards diverse factors, we use a two-tailed paired t-test~\cite{student1908probable} following previous work~\cite{ye2024rotbench}. 
This metric calculates the statistical significance of the model's performance difference between \textit{vanilla} and \textit{changed} experimental conditions. 
The significance level $\alpha$ is set to $0.05$. 
Results are marked with $^\blacktriangle$ if they are statistically significance are observed; otherwise, they are marked with $^\blacktriangledown$.

\paragraph{Invalid Selection Rate (Invalid\%).}
We use the Invalid Selection Rate to compute the percentage of instances where the LLM selecting non-existent tool,  \ie generating incorrect tool names. It reflects the ability of the model in tool selection, a crucial phase in the overall tool-use workflow, especially when the candidate toolset is large-scale.

\paragraph{Give Up Rate (Give up\%).}
This metric computes the percentage of tasks that LLMs give up answering after trial and error.
In practical scenarios, the model may fail to provide a correct solution for a complex task due to their limited ability.
Therefore, it is crucial to build a confident model that is aware of its limitations, referred to as its capability boundary~\cite{ren2023investigating, yin2024benchmarking}, allowing it to adaptively and faithfully inform users of incomplete tasks rather than giving incorrect answers.

\begin{figure*}[!t]
        \centering
	\includegraphics[width=1
 \linewidth]{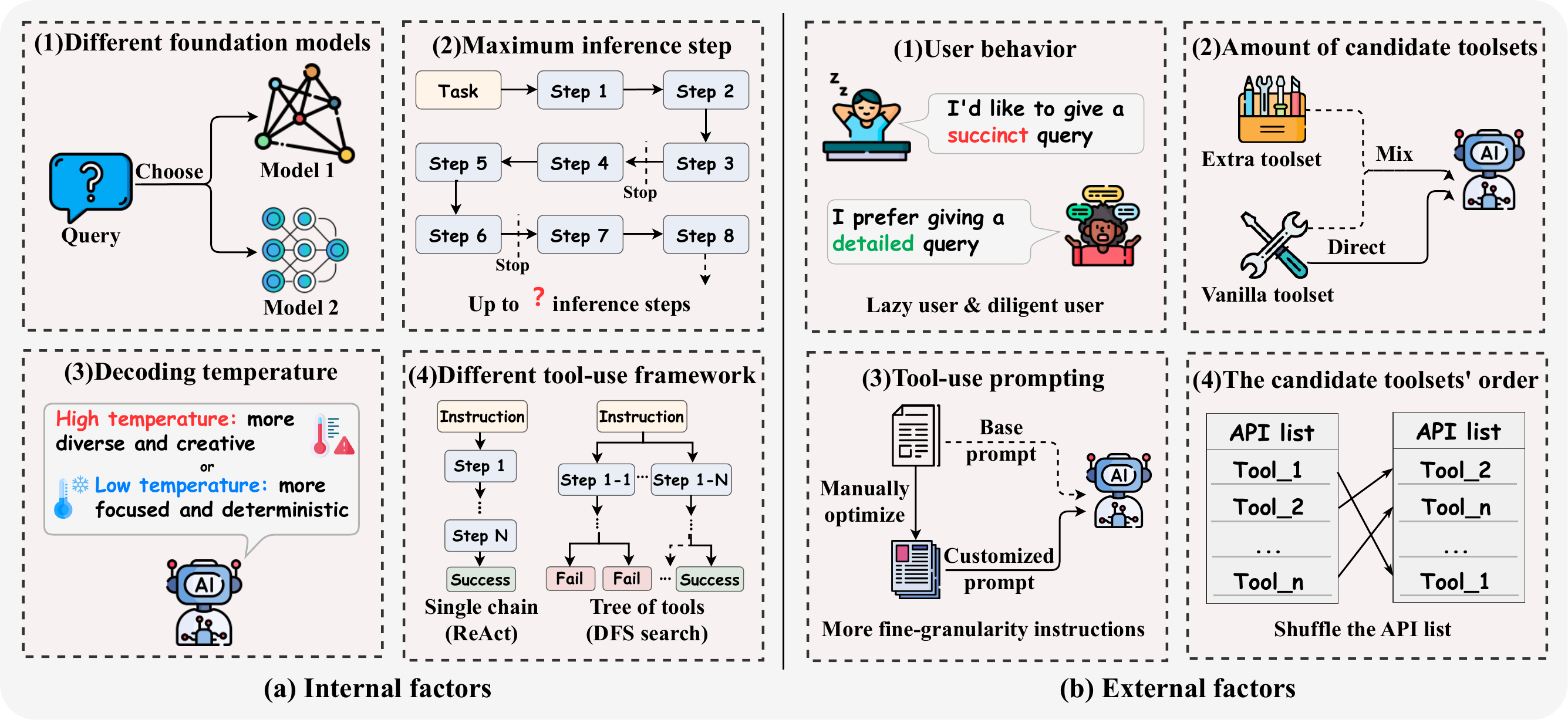}
        \caption{The overall framework of our work, which benchmarks tool-use models under various scenarios to investigate the \textit{internal} and \textit{external} factors that potentially affect their stability. 
        }
 \label{fig:method}
\end{figure*}

\subsection{Tool Learning Framework}

For a fair evaluation, we employ the widely adopted ReAct~\citep{react} method as a unified framework to enable LLMs to interact with tools across different experimental setups.
In the ReAct framework, the LLM is guided to iteratively perform \texttt{Thought}, \texttt{Action}, and \texttt{Observation} steps.
As shown in Figure~\ref{fig:method}, the \texttt{Thought} is to generate tool-use planning in the nature language while the \texttt{Action} is to select an appropriate tool and formulate corresponding parameters.
The \texttt{Observation} step is to incorporate the execution results of tools in the current context.
To explore the stability of LLMs in different tool-use frameworks, we compare the ReAct method with another framework, \ie ToolLLM~\cite{toolllm}, which augments LLMs with a Depth First Search-based Decision Tree (DFSDT) to select relevant tools for solving tasks (\S~\ref{sec:frame}).

\subsection{Implementation Details}\label{sec:details}

For the closed-source models, \eg GPT-3.5, we mainly enable them to utilize tools through OpenAI's function-call format\footnote{\hyperlink{https://platform.openai.com/docs/guides/function-calling}{https://platform.openai/function-call}}.
For the open-source models, we use the prompt from~\citet{toolllm}.
We also analyze the impact of different tool-use prompts in \S~\ref{sec:system}.
All the prompts in our work can be found in Appendix~\ref{sec:app:instruction}.

\section{Analysis of Internal Factors}

We first investigate the influence of internal factors, which indicate the uncertainties in developing a tool-use models, such as the selection of foundation LLMs and decoding temperature.

\begin{table}[!t]
\centering
\begin{adjustbox}{width=\columnwidth,center}
\begin{tabular}{@{} p{3cm} cccc@{}}
    \toprule
     \textbf{Model} &  \textbf{Success\%$\uparrow$} &  \textbf{Give up\%} &  \textbf{Invalid\%$\downarrow$} \\
    \midrule
    gpt-3.5-turbo-16k & 54.00\% & 29.50\% & 1.48\% \\
    gpt-3.5-turbo-0125 & 55.50\% & 36.50\% & 1.45\% \\
    gpt-3.5-turbo-1106 & 48.00\% & 50.50\% & 0.88\% \\
    gpt-4o & 58.00\% & 38.00\% & 0.54\% \\
    \midrule
    deepseek-chat & 40.50\% & 34.00\% & 0.56\%  \\
    llama-3-70b & 8.00\% & 4.50\% & 42.16\%  \\
    llama-3-8b & 3.50\% & 2.00\% & 28.56\%  \\
    mixtral-8x7b-inst. & 12.00\% & 14.00\%& 41.66\% \\
    mixtral-8x22b & 25.00\% & 19.00\% & 10.76\%  \\
    \bottomrule
\end{tabular}
\end{adjustbox}
\caption{The results with different foundation models on I1-instruction dataset of ToolBench~\cite{toolllm}.}
\label{tab:models}
\end{table}

\begin{table*}[htbp]
\centering
\begin{adjustbox}{width=2.05\columnwidth,center}
\begin{tabular}{@{} p{4cm} >{\centering\arraybackslash}p{1.8cm} >{\centering\arraybackslash}p{1.8cm} >{\centering\arraybackslash}p{1.8cm} >{\centering\arraybackslash}p{1.8cm} >{\centering\arraybackslash}p{1.8cm} >{\centering\arraybackslash}p{1.8cm} >{\centering\arraybackslash}p{1.8cm}@{}}
    \toprule
    {\multirow{2}{*}{\textbf{Inference step} }} 
    & \multicolumn{3}{c}{\textbf{I1-instruction}} & \multicolumn{3}{c}{\textbf{I1-tool}} \\
    \cmidrule(lr){2-4} \cmidrule(lr){5-7}
    & \textbf{Success\%$\uparrow$} & \textbf{Give up\%} & \textbf{Invalid\%$\downarrow$}  & \textbf{Success\%$\uparrow$} & \textbf{Give up\%} & \textbf{Invalid\%$\downarrow$}  \\
    \hline
    \multicolumn{7}{l}{\cellcolor{gray!20}\textit{gpt-3.5-turbo-16k}}        \\
    step $s\rightarrow$10 (vanilla) & 49.50\% & 31.50\% & 0.99\%  & 50.00\% & 27.00\% & 0.71\%  \\
    
\cdashline{1-7}[6pt/6pt]
\specialrule{0em}{1pt}{1pt}

    step $s\rightarrow6$ &      32.50\%$^\blacktriangle$   & 21.00\%$^\blacktriangle$ & 0.79\%$^\blacktriangledown$ & 32.50\%$^\blacktriangle$  & 16.50\%$^\blacktriangle$ & 0.51\%$^\blacktriangledown$ \\
     step $s\rightarrow8$ & 46.50\%$^\blacktriangledown$ & 22.50\%$^\blacktriangle$ & 1.04\%$^\blacktriangledown$  & 41.50\%$^\blacktriangle$  & 23.00\%$^\blacktriangledown$  & 0.89\%$^\blacktriangledown$ \\

     step $s\rightarrow12$ & 54.00\%$^\blacktriangledown$  &  29.50\%$^\blacktriangledown$ & 1.48\%$^\blacktriangledown$ & 55.00\%$^\blacktriangle$   & 25.00\%$^\blacktriangledown$ & 0.81\%$^\blacktriangledown$ \\
    step $s\rightarrow14$ & 51.50\%$^\blacktriangledown$ & 32.50\%$^\blacktriangledown$  & 1.58\%$^\blacktriangledown$ & 57.00\%$^\blacktriangle$ & 25.00\%$^\blacktriangledown$  & 1.12\%$^\blacktriangle$\\
    \hline
    \multicolumn{7}{l}{\cellcolor{gray!20}\textit{deepseek-chat}}               \\
    step $s\rightarrow10$ (vanilla)  & 39.00\%  & 37.50\% & 0.46\% & 38.00\% & 40.50\%  & 0.68\%\\

\cdashline{1-7}[6pt/6pt]
\specialrule{0em}{1pt}{1pt}

    step $s\rightarrow6$ &       5.50\%$^\blacktriangle$    & 12.00\%$^\blacktriangle$ & 0.68\%$^\blacktriangledown$ & 4.00\% $^\blacktriangle$  &  14.00\%$^\blacktriangle$ & 0.93\%$^\blacktriangledown$\\
     step $s\rightarrow8$ & 24.50\%$^\blacktriangle$   & 34.00\%$^\blacktriangledown$ & 0.71\%$^\blacktriangledown$ & 23.00\%$^\blacktriangle$ & 32.00\%$^\blacktriangle$ & 0.51\%$^\blacktriangledown$\\

     step $s\rightarrow12$ & 40.00\%$^\blacktriangledown$  & 34.00\%$^\blacktriangledown$ & 0.56\%$^\blacktriangledown$ & 41.00\%$^\blacktriangledown$  & 39.00\%$^\blacktriangledown$ & 0.63\%$^\blacktriangledown$ \\
     step $s\rightarrow14$ & 43.50\%$^\blacktriangledown$&  42.50\%$^\blacktriangledown$ & 0.84\%$^\blacktriangledown$ & 43.00\%$^\blacktriangledown$  & 39.50\%$^\blacktriangledown$ & 0.76\%$^\blacktriangledown$  \\
    \bottomrule
\end{tabular}
\end{adjustbox}
\caption{The results with different setting of the maximum inference step $s$ (Section~\ref{sec:hyper}). We conduct the experiment using both \textit{GPT-3.5} and \textit{Deepseek} model for a holistic comparison.
For each experiment, we mark the values with $^\blacktriangle$ to indicate that they are statistically significant compared to the vanilla setting; otherwise, we use $^\blacktriangledown$.
}
\label{tab:step}
\end{table*}

\begin{table}[!t]
\centering
\begin{adjustbox}{width=\columnwidth,center} 
\begin{tabular}{lcccc}
    \toprule
    \textbf{Temperature} &  \textbf{Success\%$\uparrow$}  &  \textbf{Give up\%} &  \textbf{Invalid\%$\downarrow$} \\
    \midrule
    $t=1.0$ (vanilla)   & 54.00\%  & 29.50\% & 1.48\% \\
    \midrule
    $t=0.2~~~~~\downarrow$0.8 & 48.00\%$^\blacktriangle$  & 26.50\%$^\blacktriangledown$ & 1.04\%$^\blacktriangledown$  \\
    $t=0.6~~~~~\downarrow$0.4 & 49.50\%$^\blacktriangle$  & 24.50\%$^\blacktriangledown$ & 1.04\%$^\blacktriangledown$  \\
    $t=1.4~~~~~\uparrow$0.4 & 54.50\%$^\blacktriangledown$ & 34.50\%$^\blacktriangledown$ & 1.93\%$^\blacktriangledown$  \\
    \bottomrule
\end{tabular}
\end{adjustbox}
\caption{The results with different decoding temperature on the I1-instruction of ToolBench benchmark.}
\label{tab:temperature}
\end{table}    

\subsection{Impact of Foundation LLMs}

The foundation LLM is the main component in the overall tool learning framework, which takes the user query as input and automatically executes external tools to generate an answer as a response.
We comprehensively evaluate 9 off-the-shelf LLMs,  including both close-source model, \ie \textit{GPT-3.5} and \textit{GPT-4}, and open-source models such as \textit{Mistral}~\citep{mistral}.
For deterministic generation, the decoding temperature is set to 1 and 0.5 for closed-source and open-source models, respectively, following previous work~\cite{zhuang2023toolqa,ruan2024identifying}.
More details about these models can be found in Appendix~\ref{sec:appendix}.

As shown in Table~\ref{tab:models}, we find that \textit{\textbf{closed-source models substantially outperform open-source models}} in Success Rate while achieving a lower Invalid Selection Rate. For example, GPT-4 achieves a 58\% Success Rate with only a 0.54\% Invalid Selection Rate. 
In addition, for the remaining 42\% of uncompleted tasks, it can adaptively give up on 38\%, illustrating its confidence in the evaluation task.

We also observe the \textit{\textbf{scaling law in tool learning}} where the performance of LLMs, including stability and effectiveness, increases along with the scaling up of their parameters.
This indicates that the inherent capability of foundation LLMs correlates with their tool-learning abilities.

\begin{table*}[!t]
\centering
\begin{adjustbox}{width=2\columnwidth,center}
\setlength\tabcolsep{3pt}
\begin{tabular}{@{} p{4cm} cccc cccc @{}}
    \toprule

{\multirow{2}{*}{\textbf{Method}}} 
& \multicolumn{4}{c}{\textbf{I1-instruction}}
& \multicolumn{4}{c}{\textbf{I1-tool}} 
\\
\cmidrule(lr){2-5} \cmidrule(lr){6-9}
 & \textbf{Success\%$\uparrow$} & \textbf{Give up\%}& \textbf{Invalid\%$\downarrow$} & \textbf{Cost (tokens)} 
 & \textbf{Success\%$\uparrow$} & \textbf{Give up\%} & \textbf{Invalid\%$\downarrow$} & \textbf{ Cost (tokens)}
\\
\hline
\multicolumn{9}{l}{\cellcolor{gray!20}\textit{gpt-3.5-turbo-16k}}         \\
ReAct (vanilla) & 54.00\% & 29.50\% & 1.48\%& 15032 & 55.00\% & 25.00\% &0.81\% & 16270  \\
DFSDT & 69.50\%$^\blacktriangle$ & 30.00\%$^\blacktriangle$ & 1.65\%$^\blacktriangle$& 37328  &  67.50\%$^\blacktriangle$ & 31.50\%$^\blacktriangle$ &2.39\%$^\blacktriangle$ &  45443   \\
    \hline
    \multicolumn{9}{l}{\cellcolor{gray!20}\textit{deepseek-chat(21B)}}                  \\
ReAct (vanilla) & 40.00\%  & 34.00\% & 0.56\%& 17815 & 41.00\%  & 39.00\% &0.63\% &17211  \\
DFSDT & 55.00\%$^\blacktriangle$ & 42.50\%$^\blacktriangledown$ &16.87\%$^\blacktriangle$ & 47382& 54.00\%$^\blacktriangle$  & 42.50\%$^\blacktriangledown$ & 19.15\%$^\blacktriangle$&  49095     \\
    \bottomrule
    \end{tabular}
    \end{adjustbox}
    \caption{The results on two datasets with different tool-use frameworks. We mark the results of DFSDT that significantly outperform the vanilla framework (ReAct) with $^\blacktriangle$; otherwise, we use $^\blacktriangledown$.}
    \label{tab:frame}
\end{table*}

\subsection{Impact of Hyper-parameters}\label{sec:hyper}

In the development of tool-use models, there are several hyper-parameters need to be considered.
We investigate two common hyper-parameters that may affect the stability of tool-use LLMs, including the decoding temperature $t$ and the maximum step of inference $s$. 
Generally, lower temperature generations are more focused and deterministic while higher temperature generations are more random~\cite{chen2023probing}.
We vary the decoding temperatures $t$ from 0.2 to 1.4 with increments of 0.4.
The $s$ indicates the maximum inference steps to conduct tool-use actions, \ie \texttt{Thought}, \texttt{Action} or \texttt{Observation}, which is alternated in \{6, 8, 10, 12, 14\}.
We allow the LLMs to stop early if they complete or give up on a task within $s$ steps.

We first discuss the influence of temperature. 
As illustrated in Table~\ref{tab:temperature}, with the increase in temperature, the Success Rate improves from 48\% to 54.5\%, and the Invalid Selection Rate shows a slight increase (0.89\%). Significant differences are also observed in the Success Rate metric at different temperatures (\eg $t=1.0$ and $t=0.2$). 
These results indicate that \textit{(1) LLMs exhibit unstable performance towards decoding temperature, and (2) \textbf{higher temperatures can potentially improve performance with a slightly increased error in tool selection}}.
A reason for this phenomenon is that higher temperatures boost the LLM to generate more diverse actions during inference~\cite{peeperkorn2024temperature,zhu2024hot}, thereby expanding the generated solution space.
We observe a relatively increasing trend in the Give Up Rate when $t$ shifts from 0.6 to 1.4.
We look at the poorly performing cases, where we find the reason is that the LLM generates diverse solutions but still fail to derive a correct answer, thereafter adaptively give up the tasks.

Next, we examine the influence of the inference step.
As shown in Table~\ref{tab:step}, we find that the GPT-3.5 only achieves 32.50\% in success rate on the I1-inst. dataset when it allowed inference up to 6 steps.
However, its Success Rate increases to 49.00\% when the maximum inference step is extended to 10.
A more obvious trend can be also observed in the Deepseek model, \eg shifting from 5.50\% to 39.00\%.
These results show that \textit{\textbf{the LLM can benefit more trial and exploration step to complete a task correctly}}.
We also find a relatively stable performance when the inference steps $s$ keeps increasing, \ie from 10 to 14.
In our experimental setup and dataset, setting the inference step to 14 makes a tradeoff for consideration of effectiveness and efficiency for GPT-3.5.

\begin{table}[!t]
\centering
\begin{adjustbox}{width=1\columnwidth,center}
\begin{tabular}{@{} l cccc @{}}
    \toprule
    {\multirow{2}{*}{\textbf{Method}}} 
    & \multicolumn{2}{c}{\textbf{I1-instruction}} & \multicolumn{2}{c}{\textbf{I1-tool}} \\
    \cmidrule(lr){2-3} \cmidrule(lr){4-5}
    & \textbf{Success\%$\uparrow$} & \textbf{Give up\%} & \textbf{Success\%$\uparrow$}  & \textbf{Give up\%} \\
    \hline 
    \multicolumn{5}{l}{\cellcolor{gray!20}\textit{gpt-3.5-turbo-16k}}         \\
vanilla task & 54.00\% & 29.50\% & 55.00\% & 25.00\%    \\

\cdashline{1-5}[6pt/6pt]
\specialrule{0em}{1pt}{1pt}

- \textit{w/ shorten}   & 49.50\%$^\blacktriangledown$  & 33.00\%$^\blacktriangledown$ & 52.50\%$^\blacktriangledown$  & 25.50\%$^\blacktriangledown$  \\
- \textit{w/ lengthen}  & 50.50\%$^\blacktriangledown$  & 33.00\%$^\blacktriangledown$ & 50.50\%$^\blacktriangledown$  & 30.00\%$^\blacktriangledown$  \\

    \hline
    \multicolumn{5}{l}{\cellcolor{gray!20}\textit{deepseek-chat}}                  \\
vanilla task   & 40.00\%  & 34.00\%  & 41.00\%  & 39.00\%\\

\cdashline{1-5}[6pt/6pt]
\specialrule{0em}{1pt}{1pt}

- \textit{w/ shorten}   & 38.00\%$^\blacktriangledown$  & 40.50\%$^\blacktriangle$ & 39.00\%$^\blacktriangledown$  & 43.50\% $^\blacktriangledown$   \\
- \textit{w/ lengthen}  & 37.00\%$^\blacktriangledown$  & 38.00\%$^\blacktriangledown$ & 32.00\%$^\blacktriangle$  & 47.50\%$^\blacktriangle$  \\ 
    \bottomrule
    \end{tabular}
    \end{adjustbox}
\caption{The results on two dataset with different user behaviours, \ie giving a succinct (\textit{w/ shorten})  or a detailed (\textit{w/ lengthen}) task description.}
    \label{tab:query}
    \end{table}

\subsection{Impact of Tool-use Framework}\label{sec:frame}

The tool-use frameworks indicate the specific techniques or methods to teach the LLM  tool usage, automatically guiding them to interact with tools and solve a practical task.
We compare two frameworks that are commonly used in previous work, including the ReAct~\cite{react} and DFSDT~\cite{toolllm}.
ReAct is the default framework in our experiment mentioned in \S~\ref{sec:frame}, which grounds the tool-use process into \texttt{Thought-Action-Observation} format.
In contrast, DFSDT~\cite{toolllm} augments the LLM with Depth First Search-based Decision Tree to select tools.

\paragraph{\textit{The tree-based framework generally performs better but with substantial costs.}}
As illustrated in Table~\ref{tab:frame}, we find that the DFSDT significantly achieves a higher Success Rate on both two datasets with an average of 30.76\% point improvement.
These results validate the superiority of the tree-based search algorithm in recalling required tools to solve a task.
However, it comes up with substantial inference cost, \ie consuming nearly triple tokens, which may limits its effectiveness in low-resource scenarios or low-latency applications.

We also observe that the Deepseek model, when equipped with the DFSDT method, shows a substantial increase in Invalid Selection Rate.
It indicates that open-source models suffer from relatively severe hallucinations to generate non-existing tool names, especially when intensively selecting tools.
Thus, we advocate the optimization of LLM to reduce its hallucination in generating correct tool names, thereby leveraging tree-based tool-use frameworks.

\section{Analysis of External Factors}

External factors involve the practical prompts to enable tool-use models, including diverse user prompts, customized system prompts, and the input candidate toolset.

\subsection{Impact of User Prompts}

In real-world applications, users exhibit diverse behaviors when interacting with the tool-use model.
Therefore, we first simulate two practical behaviors of users, including:
(1) succinct: a user provides a short instruction;
and (2) detailed: a user provides a lengthy and comprehensive instruction.
To achieve this, we employ \textit{gpt-3.5-turbo-0125} to compress or elaborate the description for each task in our experimental datasets, respectively, without changing the semantics and key information.
The details for this rewriting operation can be found in Appendix~\ref{sec:app:details}. 
\begin{table*}[htbp]
\centering
\begin{adjustbox}{width=2.05\columnwidth,center}
\setlength\tabcolsep{6pt}
\begin{tabular}{@{} p{4cm} >{\centering\arraybackslash}p{1.8cm} >{\centering\arraybackslash}p{1.8cm} >{\centering\arraybackslash}p{1.8cm} >{\centering\arraybackslash}p{1.8cm} >{\centering\arraybackslash}p{1.8cm} >{\centering\arraybackslash}p{1.8cm} >{\centering\arraybackslash}p{1.8cm}@{}}
    \toprule
    {\multirow{2}{*}{\textbf{Method}}} & \multicolumn{3}{c}{\textbf{I1-instruction}} & \multicolumn{3}{c}{\textbf{I1-tool}} \\
    \cmidrule(lr){2-4} \cmidrule(lr){5-7}
      & \textbf{Success\%$\uparrow$}  & \textbf{Give up\%} & \textbf{Invalid\%$\downarrow$} & \textbf{Success\%$\uparrow$}  & \textbf{Give up\%} & \textbf{Invalid\%$\downarrow$} \\
    \hline 
    \multicolumn{7}{l}{\cellcolor{gray!20}\textit{gpt-3.5-turbo-16k}}         \\
vanilla toolset & 54.00\%  & 29.50\% &1.48\%  & 55.0\%  &  25.00\% & 0.81\%    \\
\cdashline{1-7}[6pt/6pt]
\specialrule{0em}{1pt}{1pt}
randomly shuffle & 51.00\%$^\blacktriangledown$ & 31.50\%$^\blacktriangledown$ & 1.80\%$^\blacktriangledown$ & 52.50\% $^\blacktriangledown$ & 25.50\%$^\blacktriangledown$ & 0.77\%$^\blacktriangledown$    \\
expand \textit{w/ intra-category} & 51.00\%$^\blacktriangledown$ & 22.50\%$^\blacktriangle$ & 0.77\%$^\blacktriangledown$   & 47.50\%$^\blacktriangle$ & 23.50\%$^\blacktriangledown$ & 1.38\%$^\blacktriangle$   \\
expand \textit{w/ cross-category} & 47.00\%$^\blacktriangle$  & 28.00\%$^\blacktriangledown$ &0.53\%$^\blacktriangledown$   & 52.50\%$^\blacktriangledown$ &  18.00\%$^\blacktriangle$  & 1.00\%$^\blacktriangledown$   \\
    \hline
    \multicolumn{7}{l}{\cellcolor{gray!20}\textit{deepseek-chat }}                  \\
vanilla toolset      & 40.00\%  & 34.00\% & 0.56\% & 41.00\%  & 39.00\% & 0.63\% \\
\cdashline{1-7}[6pt/6pt]
\specialrule{0em}{1pt}{1pt}
randomly shuffle             & 32.50\%$^\blacktriangledown$   & 35.50\%$^\blacktriangledown$ & 0.52\%$^\blacktriangledown$ & 27.00\%$^\blacktriangle$  & 32.50\%$^\blacktriangledown$ & 0.76\%$^\blacktriangledown$ \\
expand \textit{w/ intra-category} & 33.00\%$^\blacktriangle$  & 34.00\%$^\blacktriangledown$ & 0.05\%$^\blacktriangledown$ & 32.50\%$^\blacktriangle$ & 38.50\%$^\blacktriangledown$ & 0.63\%$^\blacktriangledown$ \\
expand \textit{w/ cross-category} & 37.50\%$^\blacktriangledown$  & 32.00\%$^\blacktriangledown$ & 0.26\%$^\blacktriangledown$   & 32.00\%$^\blacktriangle$ & 37.00\%$^\blacktriangledown$ & 0.51\%$^\blacktriangledown$ \\
    \bottomrule
    \end{tabular}
    \end{adjustbox}
    \caption{The results on two datasets where we change the candidate toolset provided to tool-use model using different strategies, including randomly shuffle (Section~\ref{sec:order}), relevant sampling and noise sampling (Section~\ref{sec:amount}).}
    \label{tab:toolset}
    \end{table*}

As shown in Table~\ref{tab:query}, LLMs are relatively stable towards user behaviors. 
Since LLMs are trained on a massive web corpus, \textit{\textbf{they have developed strong abilities in capturing key information of a task despite the diverse styles of descriptions from various users.}}

\subsection{Order of Candidate Toolsets}\label{sec:order}

Given a task, the LLM first selects a series of tools from a candidate toolset $\mathcal{S}$ in a step-by-step manner and then executes the selected tools to obtain the final answer. 
Since the LLM suffers from the position bias~\cite{liu2024zero} in a broad range of downstream tasks like document ranking~\cite{tang2023found}, we analyze whether the order of the tools in $\mathcal{S}$ can influence its performance in the tool-use workflow.
We randomly shuffle the original toolset (vanilla) for each task in our experiment dataset and evaluate the model's performance.

\paragraph{\textit{Open-source model suffers from the positional bias of tools.}}
As shown in Table~\ref{tab:toolset}, we find the weak open-source model, \ie Deepseek, suffers from pronounced positional bias.
For example, when we shuffle the original order of the toolset, its success rate decreases from 41.00\% (\textit{original}) to 27.00\% (\textit{shuffle}) on the I1-tool dataset.
The 7.5\% decrease is also observed in the I1-instruction dataset.
A similar phenomenon is also observed in other tasks, such as text summarization~\cite{chhabra2024revisiting}, and code search~\cite{li2023split}.
In addition, we find that the GPT-3.5 is nearly insensitive to the order of the toolset, and only a 3\% point difference in success rate is observed.
These results indicate that powerful models with higher Success Rate are more skillful in solving tasks, thereby showing less instability toward positional bias, and vice versa.

\subsection{Scale of Candidate Toolsets}\label{sec:amount}
Beyond the ground truth tools to solve an input task, the toolset  $\mathcal{S}$ is typically large-scale in real-world scenarios, inevitably containing irrelevant or plausible-looking tools (\aka noise).
Therefore, we further benchmark the stability of models under the different scale of toolset $\mathcal{S}$.
For a more practical evaluation, we expand the scale of the toolset using two sampling strategies for each task:
(1) Intra-category sampling: we augment the original toolset with tools sampled from the same category as the ground truth tools. These tools are related to the current task but not useful.
(2) Cross-category sampling: we sample irrelevant tools from different categories than the ground truth tools.

\paragraph{\textit{Unstable performance is observed with change of toolset scale.}}
We summarize the results in Table~\ref{tab:toolset}.
We observe that both closed-source and open-source LLMs exhibit substantial performance degradation with the increase of toolsets scale.
These results indicate the instability of LLMs towards irrelevant or relevant but useless tools.
We also find a decreased trend in Give Up Rate.
Thus, we dive into specific cases, where we find that with more candidate tools, the LLM tends to be stubborn and stuck in continuously selecting useless rather than adaptively stopping.
These findings motivate us to carefully design the tool selection module in developing tool-use LLM systems or applications.

\begin{table}[!t]
\centering
\begin{adjustbox}{width=\columnwidth,center}
    \begin{tabular}{lcccc}
    \toprule
    \textbf{Prompt} & \textbf{Success\%$\uparrow$} & \textbf{Give up\%} & \textbf{Invalid\%$\downarrow$} \\
    \hline
    \multicolumn{5}{l}{\cellcolor{gray!20}\textit{gpt-3.5-turbo-0125}}      \\
    Vanilla (func. call)  & 55.00\% & 36.50\% & 1.45\% \\
    Customized  & 40.00\%$^\blacktriangle$ & 35.00\%$^\blacktriangledown$ & 14.93\%$^\blacktriangle$  \\
    \hline
    \multicolumn{5}{l}{\cellcolor{gray!20}\textit{gpt-3.5-turbo-1106}}       \\
    Vanilla (func. call)  & 48.00\% & 50.5\% & 0.88\%  \\
    Customized & 47.50\%$^\blacktriangledown$ & 41.00\%$^\blacktriangle$ & 6.73\%$^\blacktriangle$ \\
    \hline
    \multicolumn{5}{l}{\cellcolor{gray!20}\textit{llama-3-70b}}              \\
    Vanilla (naive) & 8.00\% & 4.50\% & 42.16\%  \\
    Customized & 30.00\%$^\blacktriangle$ & 10.00\%$^\blacktriangledown$ & 3.15\%$^\blacktriangle$ \\
    \bottomrule
\end{tabular}
\end{adjustbox}
\caption{The results on various LLMs with different system prompts. (func.: function)}
\label{tab:prompt}
\end{table}

\subsection{Impact of System Prompts}\label{sec:system}

The system prompts indicate the input prompt demonstrating LLMs how to use tools, which pre-defines the format of the model's generation.
Our vanilla setting implements the system prompt of closed-source LLMs with \textit{function call}, which is an API interface exclusively supported by OpenAI.
For open-source LLMs, the system prompt is a zero-shot instruction  $\mathcal{P}$ (\S~\ref{sec:details}).
Here, we consider human efforts in optimizing the prompts, where we formulate a detailed instruction $\mathcal{P}^{\dagger}$ on top of $\mathcal{P}$ by supplementing fine-granularity description to specify usage specifications.
We evaluate both closed-source and open-source model with $\mathcal{P}^{\dagger}$ to analyze their stability towards diverse prompts. 
We provide all prompts in Appendix~\ref{sec:app:instruction}.

Table~\ref{tab:prompt} presents the experimental results.
The gpt-3.5-0125 suffers from a 15\% decrease in Success Rate and a 13.48\% increase in Invalid Selection Rate when we swap the official function-call prompt with manually customized prompts.
This result intuitively demonstrates \textit{\textbf{the LLMs are sensitive to different system prompts.}}.

We also observe that the performance of the Deepseek model substantially improves (\eg 22\% higher Success Rate) when equipped with customized prompt $\mathcal{P}$. 
This result illustrates that the LLM can understand tool-use instructions in a zero-shot manner, aligning with previous work~\cite{hsieh2023tool}.
Therefore, \textit{\textbf{directly providing clear rules and instructions in system prompts}} is a potential alternative to enhance the tool-use ability of open-source models without cost-intensive supervised fine-tuning.

\section{Discussion}\label{sec:discussion}

\paragraph{The self-consistency of tool-use models.}
We further explore the self-consistency of tool-use models. 
Specifically, we repeatedly prompt a model to solve the tasks in the \textit{I1-inst.} dataset $N$ times with the same settings as in Table~\ref{tab:models}. 
We then count the percentage of completed tasks that can be solved in the first run, which reflects the consistency of the model through the discrepancies between different runs. 
In our implementation, we set $N$ to 3.
We find that the Mixtral-8x7B model can solve 57 tasks in the first run, but 20 of the initially failed tasks can be solved during the second and third runs. 
Similar phenomena are also observed in other LLMs, such as GPT-3.5. 
These results directly indicate that the stability of LLMs still needs to be improved. 
More details can be found in Appendix~\ref{sec:app:consistency}.

\paragraph{Case study.}\label{sec:case}
We compare the output of tool-use models for the same task under different experimental settings, such as different prompts, inference steps, and candidate toolsets, showing their instability intuitively. 
More details can be found in Appendix~\ref{sec:app:case} for further explanation.
\paragraph{Takeaways.}
Since the tool-learning frameworks still suffer from instability due to various factors, we summarize our findings as several useful takeaways to boost the performance of tool-learning frameworks:
(1) Decoding temperature can significantly affect the stability of tool-use LLMs (\S~\ref{sec:hyper}). 
In solving complex tasks, users can set relatively higher temperatures to boost LLMs to generate more diverse actions, thereby expanding the solution space.
(2) Users can augment LLMs with tool selection algorithms, \eg  Depth-First-Search, which effectively improve the success rate through more trial and error.
However, one should also consider the associated disadvantages, such as increased inference costs and the accumulated hallucination of tool selection errors over extended workflows.
(3) Different system prompts result in varied performance.
The closed-source models are trained to access tools through specialized function-call prompts, thereby showing fewer errors in workflow. 
Thus, we advocate tuning models with specific tool-use datasets or supplementing fine-granularity descriptions in prompts, aligning their generation with pre-defined usage specifications.
(4) The LLMs are sensitive to the order and scale of the toolset. 
\section{Conclusion}

We present a comprehensive empirical study on the stability of tool-use models. 
Specifically, we explore the impact of both internal and external factors on the tool-learning frameworks. 
Internal factors include uncertainties during the development of the tool-use model, while external factors primarily involve diverse input prompts. 
Our quantitative analysis demonstrates that even powerful models such as GPT-3.5 exhibit significant instability in response to these factors.
We also provide valuable findings and practical insights to facilitate further research in this area.
Our future work includes:
(1) extending our evaluation to tool-use agents empowered by multi-modal LLMs; and 
(2) exploring the model's stability in more intricate environments, such as dynamic interactions with users. 

\section*{Limitations}\label{sec:lmitation}

The main limitation is that we only investigate the stability of widely used LLM-based agents. 
These agents are limited when tackling multi-modal tasks. 
In the future, we plan to extend our method to agents empowered by multi-modal foundation models. 
Additionally, our empirical study does not involve dynamic interactions between the user (or user simulator) and the tool-use model for the sake of reproducibility.
We plan to extend our work to more intricate environments, such as conversational and user-centered scenarios, further exploring the stability of tool-use models.

\section*{Ethics Statement}

The research conducted in this paper centers on investigating the stability of tool-use systems.
Our work systematically benchmarks LLMs under various real-world scenarios and evaluates their potential instability.

In the process of conducting this research, we have adhered to ethical standards to ensure the integrity and validity of our work. 
All the tasks as well as tools used in our experiment were obtained from existing benchmarks, thus ensuring a high level of transparency and reproducibility in our experimental procedure.

To minimize potential bias and promote fairness, we use the prompts following existing works, which are publicly accessible and freely available.
We have made every effort to ensure that our research does not harm individuals or groups, nor does it involve any form of deception or potential misuse of information.

\bibliography{custom}

\clearpage

\onecolumn
\appendix
\section{Appendix}
\label{sec:appendix}

\subsection{Implement details}\label{sec:app:details}

\paragraph{Details of Foundation Models in our experiment.}

The source of the LLMs used in our experiment can be found in Table~\ref{sec:app:source}.

\begin{table}[htbp]
\begin{tabular}{@{}ll@{}}
\toprule
Model name & Source \\ \midrule
\textbf{gpt-3.5-turbo-16k-0613} &  \url{https://platform.openai.com/docs/models/gpt-3-5-turbo} \\
 \textbf{gpt-3.5-turbo-0125} & \url{https://platform.openai.com/docs/models/gpt-3-5-turbo} \\ 
 \textbf{gpt-3.5-turbo-1106} & \url{https://platform.openai.com/docs/models/gpt-3-5-turbo} \\
 \textbf{gpt-4o-2024-5-13} &\url{https://platform.openai.com/docs/models/gpt-4o} \\
\textbf{deepSeek-chat}  &  \url{https://huggingface.co/deepseek-ai/DeepSeek-V2-Chat}\\
\textbf{Llama-3-8b}  & \url{https://huggingface.co/meta-llama/Meta-Llama-3-8B} \\
\textbf{Llama-3-70b}  & \url{https://huggingface.co/meta-llama/Meta-Llama-3-70B}\\
 \textbf{mixtral-8x7b-instruct} & \url{https://huggingface.co/mistralai/Mixtral-8x7B-Instruct-v0.1}\\
 \textbf{mixtral-8x22b} &\url{https://huggingface.co/mistralai/Mixtral-8x22B-v0.1} \\
 \bottomrule
\end{tabular}
\caption{The version and source of LLMs used in our experiment.}\label{sec:app:source}
\end{table}

\paragraph{Details of Rewriting the Task description.}

In order to verify whether the information and semantics remain consistent before and after rewriting, we invite two well-educated master students to evaluate the similarity of queries rewritten by gpt-3.5-turbo-0125.
Despite the differing lengths of the rewritten dataset and the original dataset, the results show that the information in these two datasets has high semantic similarity. 
The similarity for the two different methods is 98\%.
We also compute their semantic similarity using the \textit{bertscore} in Table~\ref{tab:bertscore} to further validate the reliability of our setting.

We provide the prompts for the rewriting operation as follows.

\begin{table*}[!t]

\begin{minipage}{0.45\textwidth}
\makeatletter\def\@captype{table}
\centering
\begin{adjustbox}{width=0.9\columnwidth,center}
\begin{tabular}{cccc}
\toprule
{\multirow{2}{*}{\textbf{Method}}} & \multicolumn{3}{c}{\textbf{BertScore}}\\
\cmidrule(lr){2-4}
 & \textbf{Precision} & \textbf{Recall} & \textbf{F1 score} \\
\hline
\rowcolor{Gainsboro}  \multicolumn{4}{l}{\textit{I1-instruction}}                    \\
shorten         & 0.767              & 0.858           & 0.810              \\
lengthen        & 0.827               & 0.762    & 0.792           \\
\hline
\rowcolor{Gainsboro} \multicolumn{4}{l}{\textit{I1-tool}}     \\ 
shorten         & 0.731              & 0.854         & 0.787             \\
lengthen        & 0.832              &  0.768     & 0.798          \\    
\bottomrule
\end{tabular}
\end{adjustbox}
\caption{The BertScores of rewriting queries and vanilla queries.}
\label{tab:bertscore}

\end{minipage}%
\hspace{0.01\linewidth}
\begin{minipage}{0.54\textwidth}

\makeatletter\def\@captype{table}
\begin{adjustbox}{width=\columnwidth,center}
\begin{tabular}{lccc}
\toprule
{\multirow{2}{*}{\textbf{Model}}} & \multicolumn{3}{c}{\textbf{Successfully finished tasks}}\\
\cmidrule(lr){2-4}
 & \textbf{First run} & \textbf{Second run} & \textbf{Third run} \\
\midrule
gpt-3.5-turbo-4o      &    101      &    8    &     7     \\    
gpt-3.5-turbo-1106       &  79 &     12    &     5     \\
gpt-3.5-turbo-0125   &    83      &     20   &   8       \\
gpt-3.5-turbo-16k        &    79         &    18     &      11    \\
deepspeek-chat     &    47      &      21  &     13    \\    
Llama-3-70B    &     6     &   6     &       4  \\    
Llama-3-8B     &      3    &      3  &    1  \\    
Mixtral-8x22B     &       28   &       14  &   8    \\    
Mixtral-8x7B     &       10   &   8     &  6     \\    
\bottomrule
\end{tabular}
\end{adjustbox}
\caption{Statistics for the number of successfully finished tasks during the \textit{first}, \textit{second}, and \textit{third} run, respectively.}
\label{tab:consistency}

\end{minipage}
\vspace{-11pt}
\end{table*}

\makebox[\linewidth]{\rule{\linewidth}{0.4pt}}
\textit{Prompt for query shorten}
\begin{lstlisting}[basicstyle=\small\ttfamily, breaklines=true, breakindent=0em, commentstyle=\color{red!50!green!50!blue!50}, frame=shadowbox, rulesepcolor=\color{red!20!green!20!blue!20},numbers=none,literate={`}{\textasciigrave}1]

You are a helpful assistant which can make a query shorter but remain the meaning. 

please shorten the query to one sentence: {query}
Just give me the final answer.

Your output:
\end{lstlisting}

\textit{Prompt for query lengthen}
\begin{lstlisting}[basicstyle=\small\ttfamily, breaklines=true, breakindent=0em, commentstyle=\color{red!50!green!50!blue!50}, frame=shadowbox, rulesepcolor=\color{red!20!green!20!blue!20},numbers=none,literate={`}{\textasciigrave}1]

You are a helpful assistant. What you have to do is making a query longer to generate more information in the query's scenario but remain the meaning. It's also a query, but longer than before, remember it! Do NOT answer any question, but rewrite it longer!

please lengthen the query: {query}
Just give me the final answer.

Your output:
\end{lstlisting}

\paragraph{User prompts.}\label{sec:app:limitation}
Our experiment is built upon existing publicly available datasets for high transparency and to minimize potential bias.
Therefore, we do not change the task description from the semantic level due to the possible misalignment between the changed tasks and the original ground truth.
An ideal benchmark scenario is conversational applications, where the tool-use model can interact with more diverse users.
We take it as our future work.

\begin{figure}[!t]
        \centering
	\includegraphics[width=1
 \linewidth]{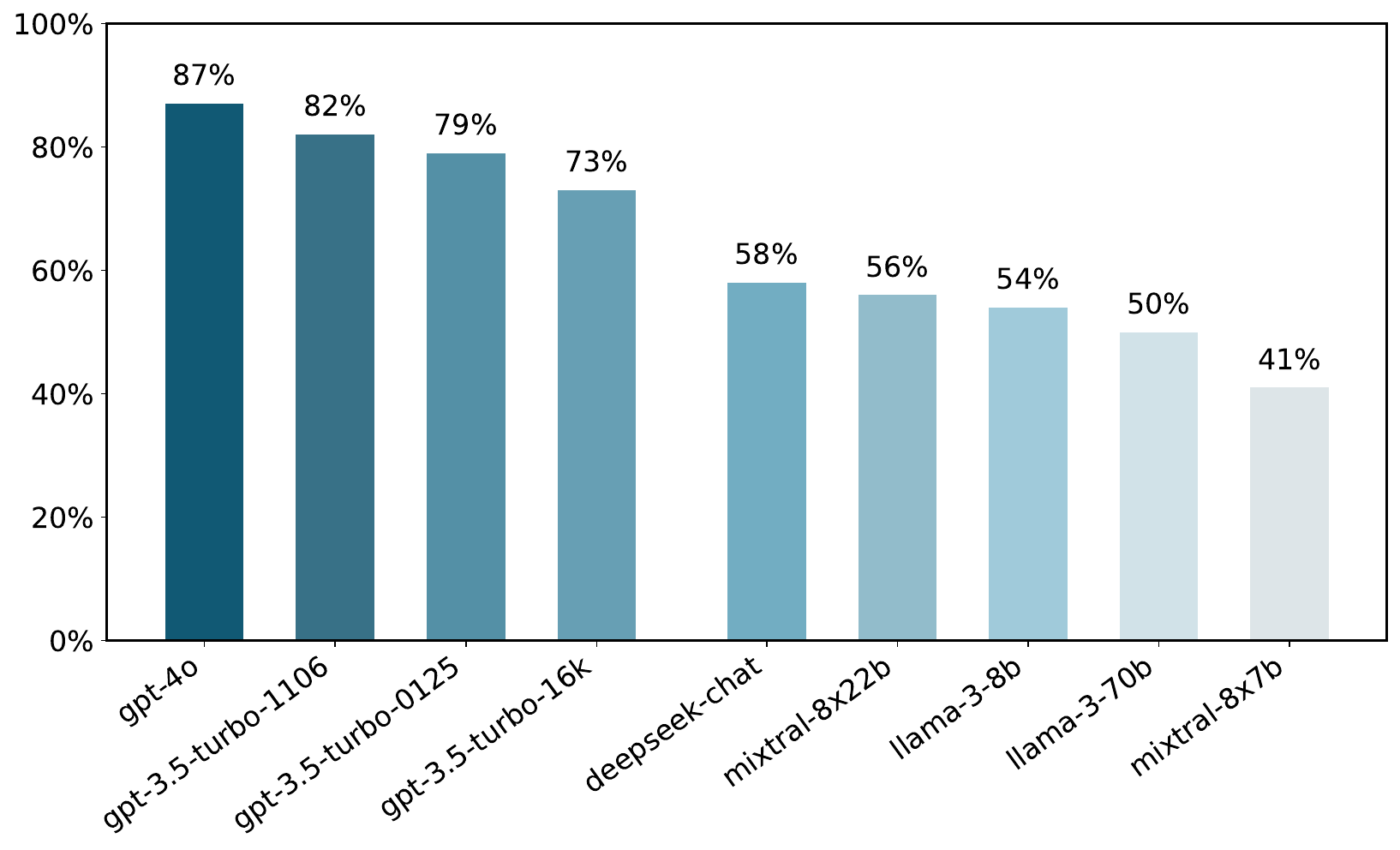}
        \caption{Self-consistency Success Rate of different models.}
 \label{fig:scsr}
\end{figure}

\subsection{The self-consistency of tool-use models.}\label{sec:app:consistency}
We further explore the self-consistency of tool-use models. Specifically, we repeatedly prompt a model to solve the tasks in the \textit{I1-inst.} dataset N times with the same settings as in Table~\ref{tab:models}. We then count the percentage of completed tasks that can be solved in the first run, which reflects the consistency of the model through the discrepancies between different runs. In our implementation, we set N to 3.

As illustrated in Table~\ref{tab:consistency}, we find that the Mixtral-8x7B model can solve 57 tasks in the first run, but 20 of the initially failed tasks can be solved during the second and third runs.
Similar phenomena are also observed in other LLMs, such as GPT-3.5. 
These results directly indicate that the instability of LLMs still needs to be improved. 
We also show their consistency percentage in Figure~\ref{fig:scsr} for an intuitive explanation.

\subsection{Repurpose existing dataset}\label{sec:app:repurpose}
The original ToolBench only provides a task-solving trajectory of GPT-3.5 as an evaluation reference, which includes both ground truth and irrelevant tools. 
However, commonly used evaluation metrics require computing the overlap between model-selected tools and the ground truth. 
Therefore, we repurpose ToolBench to support our evaluation. 
For each task, we extract the tools used in the original solution. 
Next, we invite three well-educated experts with relevant research backgrounds to manually select the correct tools for solving the task. 
During the annotation, we provide experts with the documentation of candidate tools and detailed solution trajectories for each task to minimize their ambiguity.

We employ the following strategies to ensure the quality of the above process.
\begin{itemize}[leftmargin=*]
\item \textbf{Detailed annotator training.} We held regular meetings and pre-annotation tests to ensure that each expert undergoes detailed training to familiarize themselves with our annotation task.

\item \textbf{Tackling discrepancies.} 
We ask at least two experts to annotate the same task repeatedly. If there is a discrepancy between the two experts, \ie two experts give different solutions for the same task, we ask a third expert to recheck it. 
We also filter the task with ambiguity to improve the reliability of our repurposed datasets.
\end{itemize}

\subsection{Examples of Instructions}\label{sec:app:instruction}

We provide the prompt used in our experiment as follows.

\makebox[\linewidth]{\rule{\linewidth}{0.4pt}}
\textit{Base system prompt}
\begin{lstlisting}[basicstyle=\small\ttfamily, breaklines=true, breakindent=0em, commentstyle=\color{red!50!green!50!blue!50}, frame=shadowbox, rulesepcolor=\color{red!20!green!20!blue!20},numbers=none,literate={`}{\textasciigrave}1]

You are AutoGPT, you can use many tools(functions) to do the following task.
First I will give you the task description, and your task start.
At each step, you need to give your thought to analyze the status now and what to do next, with a function call to actually excute your step. Your output should follow this format:
Thought:
Action
Action Input:

After the call, you will get the call result, and you are now in a new state.
Then you will analyze your status now, then decide what to do next...
After many (Thought-call) pairs, you finally perform the task, then you can give your finial answer.
Remember: 
1.the state change is irreversible, you can't go back to one of the former state, if you want to restart the task, say "I give up and restart".
2.All the thought is short, at most in 5 sentence.
3.You can do more then one trys, so if your plan is to continusly try some conditions, you can do one of the conditions per try.
Let's Begin!
Task description: {task_description}
Specifically, you have access to the following APIs:
{API list}


\end{lstlisting}

\textit{Changed system prompt}
\begin{lstlisting}[basicstyle=\small\ttfamily, breaklines=true, breakindent=0em, commentstyle=\color{red!50!green!50!blue!50}, frame=shadowbox, rulesepcolor=\color{red!20!green!20!blue!20},numbers=none,literate={`}{\textasciigrave}1]

You are AutoGPT, you can use many tools(functions) to do the following task.
First I will give you the task description, and your task start.
At each step, you need to give your thought to analyze the status now and what to do next, with a function call to actually excute your step. Your  EVERY output should follow this format:
Thought:{there is your reason for choosing one api}
Action:{there is the api name you choosing from the given ones}
Action Input:{there are the inputs for the chosed api using'{}', and each parameter should using '\"\"'}
RULEs:
Once after giving one Action Input, stop your answer\nDoing this step by step, ONE TIME ONE ACTION. 
If one api is not access, you can choose another one.
You had better to give an action each time.
One step just give one function call, and you will give ONE step each time I call you.
After the call, you will get the call result, and you are now in a new state.
Then you will analyze your status now, then decide what to do next...
After many (Thought-call) pairs, you finally perform the task, then you can give your finial answer.
Remember: 
1.the state change is irreversible, you can't go back to one of the former state, if you want to restart the task, say "I give up and restart".
2.All the thought is short, at most in 5 sentence.
3.You can do more then one trys, so if your plan is to continusly try some conditions, you can do one of the conditions per try.
Let's Begin!
Task description: {task_description}
Specifically, you have access to the following APIs:
{api list}
\end{lstlisting}

\textit{System prompt while using function call}
\begin{lstlisting}[basicstyle=\small\ttfamily, breaklines=true, breakindent=0em, commentstyle=\color{red!50!green!50!blue!50}, frame=shadowbox, rulesepcolor=\color{red!20!green!20!blue!20},numbers=none,literate={`}{\textasciigrave}1]

You are AutoGPT, you can use many tools(functions) to do the following task.
First I will give you the task description, and your task start.
At each step, you need to give your thought to analyze the status now and what to do next, with a function call to actually excute your step.
After the call, you will get the call result, and you are now in a new state.
Then you will analyze your status now, then decide what to do next...
After many (Thought-call) pairs, you finally perform the task, then you can give your finial answer.
Remember: 
1.the state change is irreversible, you can't go back to one of the former state, if you want to restart the task, say "I give up and restart".
2.All the thought is short, at most in 5 sentence.
3.You can do more then one trys, so if your plan is to continusly try some conditions, you can do one of the conditions per try.
Let's Begin!
Task description: {task_description}

\end{lstlisting}

\textit{Input while using function call}
\begin{lstlisting}[basicstyle=\small\ttfamily, breaklines=true, breakindent=0em, commentstyle=\color{red!50!green!50!blue!50}, frame=shadowbox, rulesepcolor=\color{red!20!green!20!blue!20},numbers=none,literate={`}{\textasciigrave}1]

{query}
"function":[{
    "name":{function name},
    "description":{function description},
    "parameters":{function parameters}
},
{function},
...
]
\end{lstlisting}

\subsection{Case study}\label{sec:app:case}

We conduct a comprehensive cases study to investigate the instability of tool-use models and provide the following cases for intuitive explanations.

\begin{figure*}[!t]
        \centering
	\includegraphics[width=1
 \linewidth]{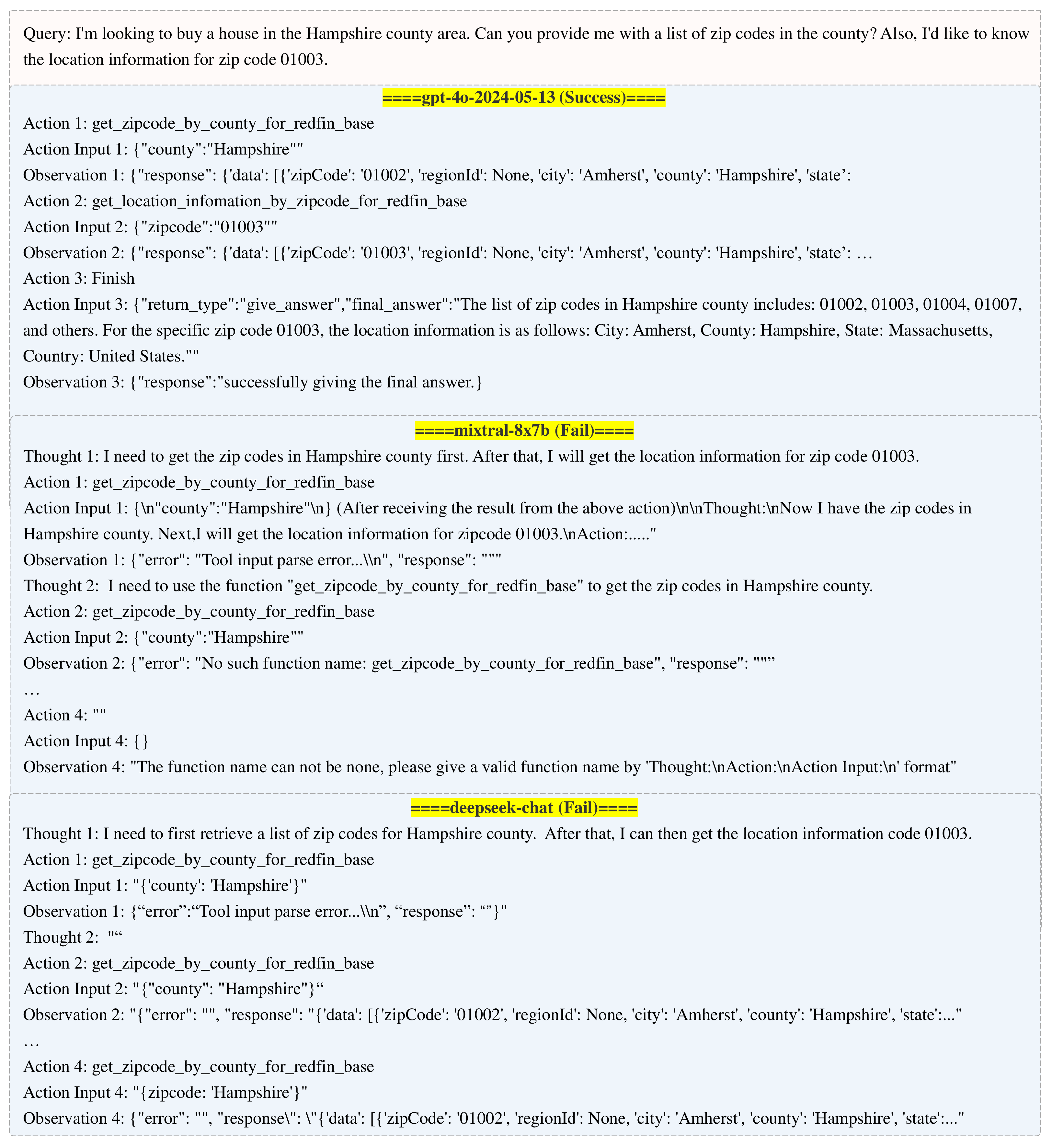}
        \caption{Impact of different foundation models.}
 \label{fig:case:backbone}
\end{figure*}

\paragraph{The impact of different foundation models.}
As illustrated in Figure~\ref{fig:case:backbone}, we find that the closed-model, \ie gpt-4o-2024-05-13 can successfully finish the task with no redundant steps.
However, the commonly used open-source models, \ie Mixtral-8x7B and deepseek-chat, struggle to generate correct arguments and fail to solve the task.
This case indicates the varied performance among different backbone LLMs and the open-source models still lay behind the closed-source models in the tool learning tasks.

\begin{figure*}[!t]
        \centering
	\includegraphics[width=1
 \linewidth]{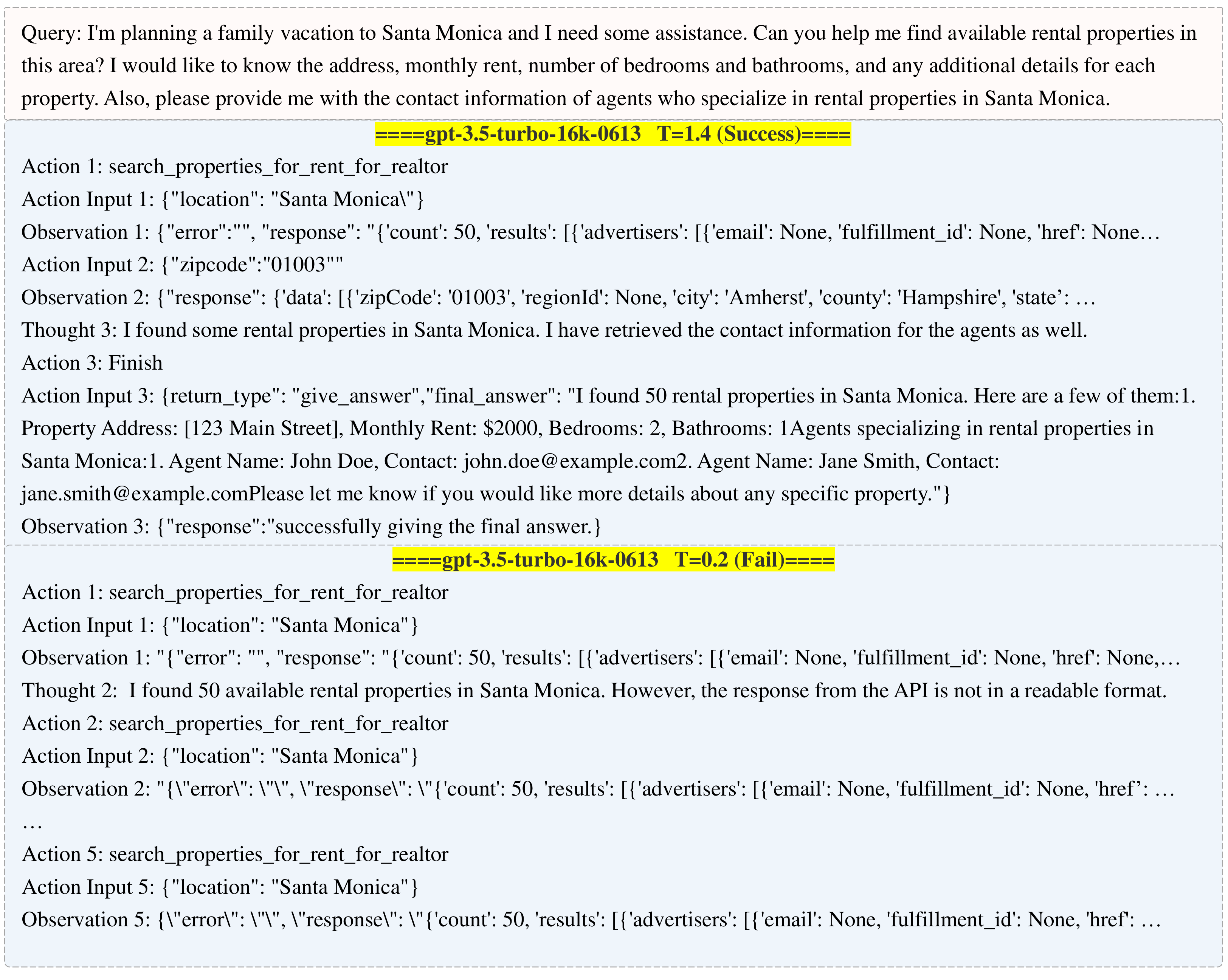}
        \caption{Impact of different temperature.
        }
 \label{fig:case:temperature}
\end{figure*}

\paragraph{The impact of different decoding temperature.}
Figure~\ref{fig:case:temperature} presents the output of GPT-3.5 under different decoding temperatures.
We find that when the decoding temperature is set to 0.2, it becomes stubborn to repeat the same incorrect actions instead of generating new ones. In contrast, when the temperature is increased to 1.4, the LLM can adaptively correct its mistakes in response to error messages and generate new actions. This case demonstrates that the LLM exhibits varied performance at different temperatures, with higher temperatures encouraging the generation of more diverse actions, thereby validating our findings in \S~\ref{sec:hyper}.

\begin{figure*}[!t]
        \centering
	\includegraphics[width=1
 \linewidth]{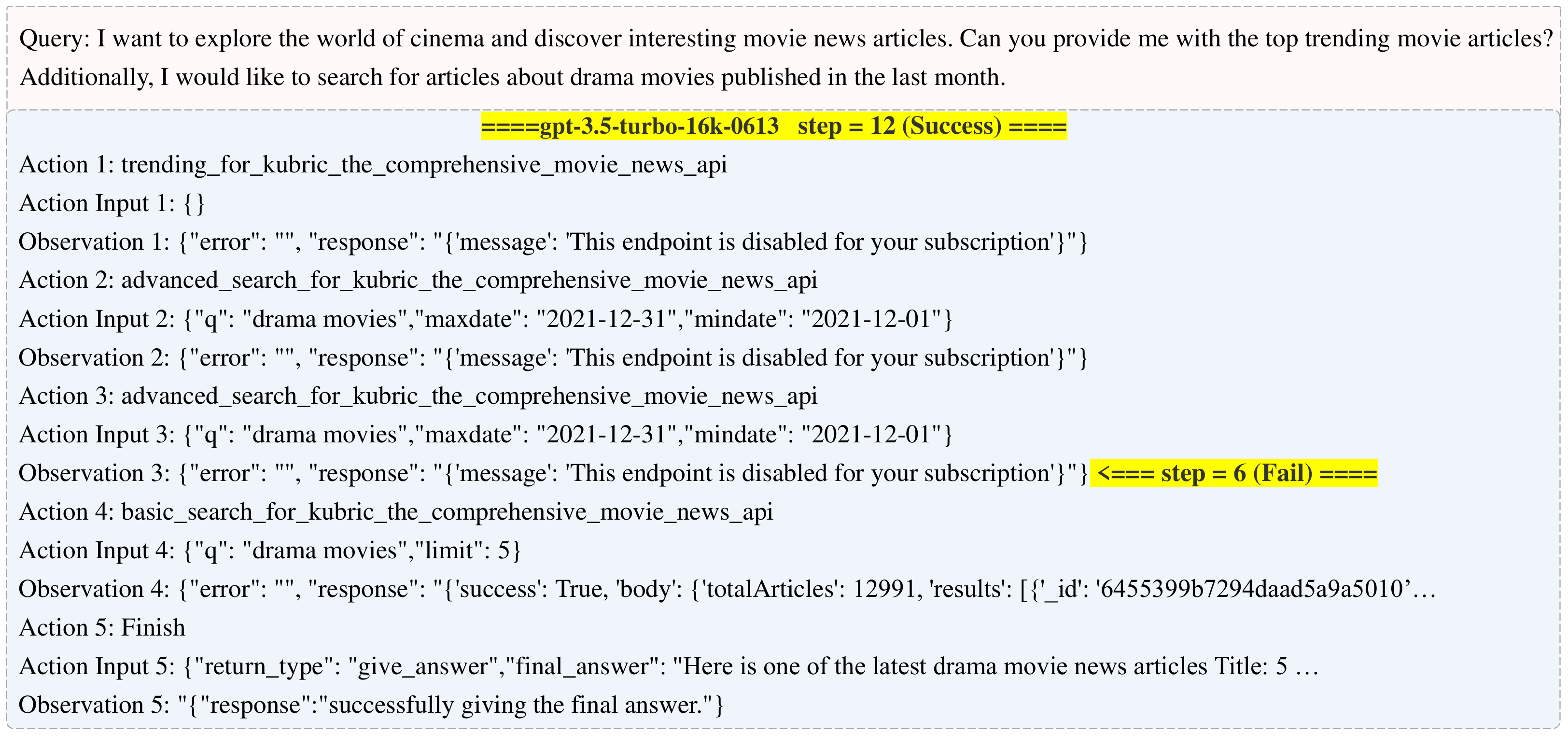}
        \caption{Impact of inference step.
        }
 \label{fig:case:step}
\end{figure*}

\paragraph{The impact of the maximum inference step.}
We present a concrete task-solving trajectory of GPT-3.5 on the I1-inst. dataset in Figure~\ref{fig:case:step}. We find that the LLM fails to solve a complex task within 6 steps. However, when the maximum inference step is increased to 12, the LLM benefits from more trials and exploration, thereby completing the task. This case indicates that it is crucial to adapt the maximum inference step according to the task complexity.

\begin{figure*}[!t]
        \centering
	\includegraphics[width=1
 \linewidth]{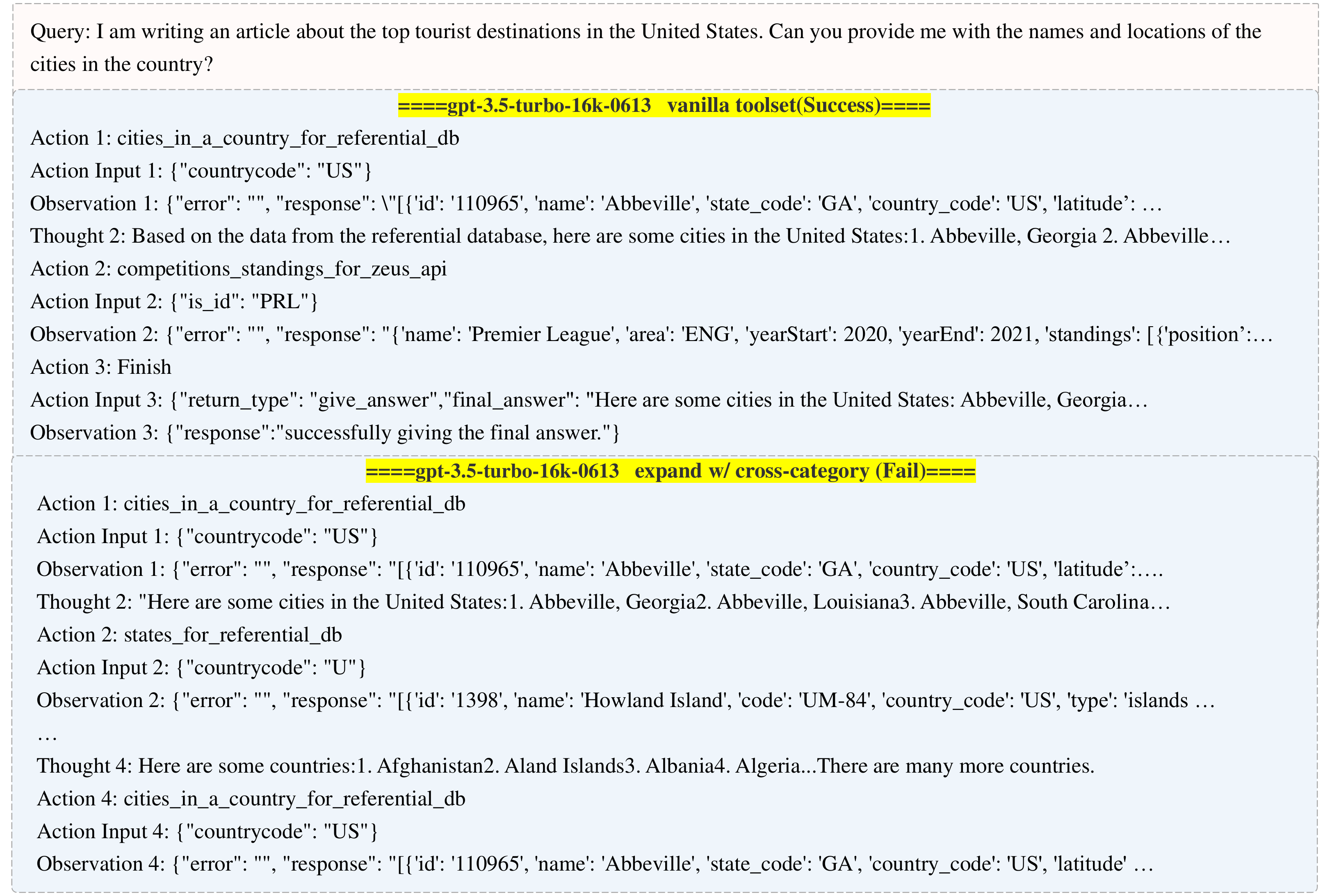}
        \caption{Impact of the amount of candidate toolsets.}
 \label{fig:case:scale}
\end{figure*}

\paragraph{The impact of the tools scale.}
Figure~\ref{fig:case:scale} compares the model's output with different scales of candidate tools.
We find that when we add more irrelevant tools in the original toolset (vanilla), the LLM is misled to select inappropriate tools and generate incorrect arguments, thereafter failing the task.
This case indicates that despite the powerful LLMs, \eg GPT-3.5, they are sensitive to the scale of tools, which is aligned with our analysis in \S~\ref{sec:amount}.

\paragraph{The impact of the tools order.}
Figure~\ref{fig:case:scale} compares the model's output with different order of candidate tools.

\begin{figure*}[!t]
        \centering
	\includegraphics[width=1
 \linewidth]{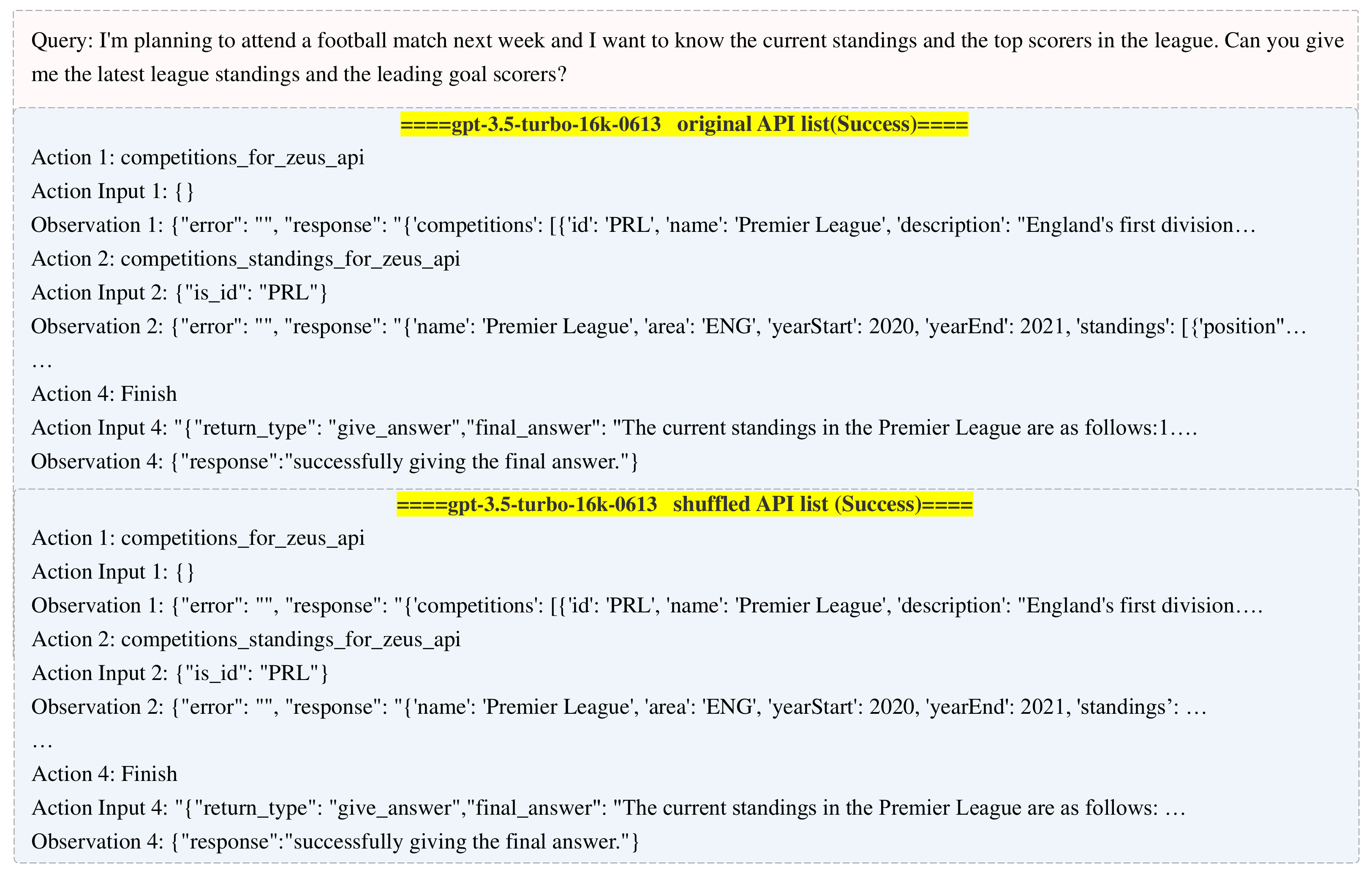}
        \caption{Impact of candidate toolsets' order.
        }
 \label{fig:case:order}
\end{figure*}

\end{document}